%% file: main.tex
\documentclass[10pt,twocolumn,letterpaper,dvipsnames]{article}
\usepackage{tikz}
\usepackage[pagenumbers]{wacv} 
\usepackage{multirow}
\usepackage{array}

\definecolor{wacvblue}{rgb}{0.21,0.49,0.74}
\usepackage[pagebackref,breaklinks,colorlinks,allcolors=wacvblue]{hyperref}

\def\wacvPaperID{1435} 
\def\confName{WACV}
\def\confYear{2026}

\title{Enhancing Monocular 3D Hand Reconstruction with Learned Texture Priors}

\author{Giorgos Karvounas$^{1,3}$\\
{\tt\small gkarv@ics.forth.gr}
\and
Nikolaos Kyriazis$^{1}$\\
{\tt\small kyriazis@ics.forth.gr}
\and
Iason Oikonomidis$^{1}$\\
{\tt\small oikonom@ics.forth.gr}
\and
Georgios Pavlakos$^{2}$\\
{\tt\small pavlakos@cs.utexas.edu}
\and
Antonis A. Argyros$^{1,3}$\\
{\tt\small argyros@ics.forth.gr }
\and
$^{1}$ICS-FORTH, $^{2}$University of Texas at Austin, $^{3}$University of Crete
}

\begin{document}
\twocolumn[{
\renewcommand\twocolumn[1][]{#1}
\maketitle

\begin{center}
    \vspace{-0.26in}
    \centerline{
    \includegraphics[width=.9\linewidth]{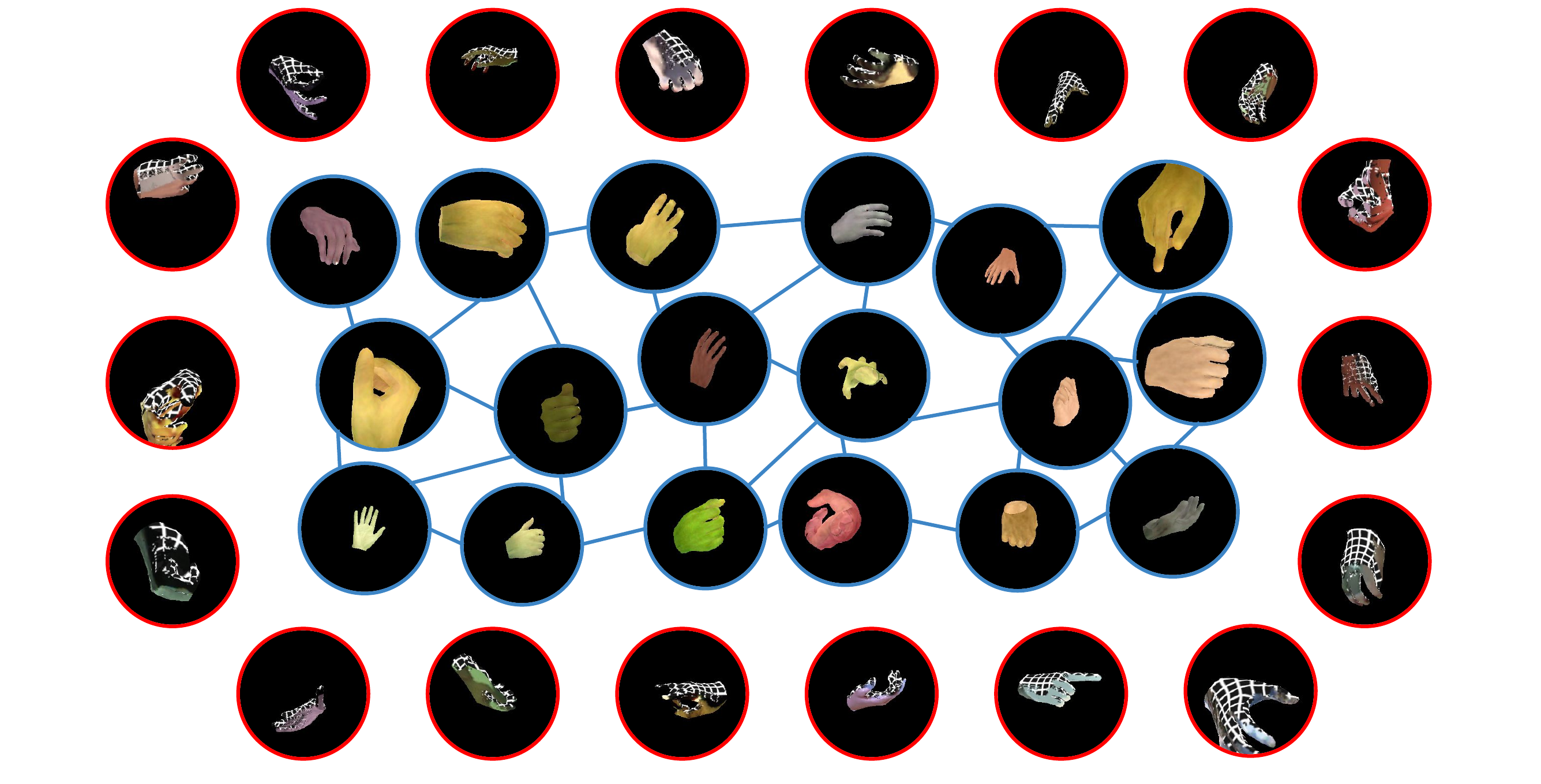}
     }
    \vspace{-0.08in}
   \captionof{figure}{
   \textit{In-the-wild} hand images provide the richest, but most incomplete and noisy input for building texture priors (samples outlined \textcolor{red}{red}, grid regions indicate missing information in the monocular setting). Our method is able to consolidate such data, enabling robust full texture suggestions (samples outlined \textcolor{blue}{blue}).
   }
   \vspace{-0.02in}
\label{fig:teaser}
\end{center}
}]

\input{sec/0_abstract}    
\input{sec/1_intro}
\input{sec/2_related}
\input{sec/3_method}

\input{sec/4_results}
\input{sec/5_conclusion}

\section*{Acknowledgments} 
This work was co-funded by the European Union (EUHE Magician - Grant Agreement 101120731) and the internal FORTH-ICS project ``Large-scale, Diverse 3D Modeling of Human Hands''- DiveHands. The authors also gratefully acknowledge the support for this research from the VMware University Research Fund (VMURF). 
{
    \small
    \bibliographystyle{ieeenat_fullname}
    \bibliography{main}
}

\end{document}

%% file: sec/0_abstract.tex
\begin{abstract}

We revisit the role of texture in monocular 3D hand reconstruction, not as an afterthought for photorealism, but as a dense, spatially grounded cue that can actively support pose and shape estimation. Our observation is simple: even in high-performing models, the overlay between predicted hand geometry and image appearance is often imperfect, suggesting that texture alignment may be an underused supervisory signal. We propose a lightweight texture module that embeds per-pixel observations into UV texture space and enables a novel dense alignment loss between predicted and observed hand appearances. Our approach assumes access to a differentiable rendering pipeline and a model that maps images to 3D hand meshes with known topology, allowing us to back-project a textured hand onto the image and perform pixel-based alignment. The module is self-contained and easily pluggable into existing reconstruction pipelines. To isolate and highlight the value of texture-guided supervision, we augment HaMeR~\cite{pavlakos2024reconstructing}, a high-performing yet unadorned transformer architecture for 3D hand pose estimation. The resulting system improves both accuracy and realism, demonstrating the value of appearance-guided alignment in hand reconstruction.

\end{abstract}

%% file: sec/1_intro.tex
\section{Introduction}
\label{sec:intro}



Reconstructing 3D hand pose and shape from a single image is a long-standing scientific challenge. Despite its underconstrained nature, accurate monocular estimation is critical across domains. In healthcare, it supports remote physiotherapy and motor assessment. In robotics, it enables teleoperation and human-robot collaboration. In education, it powers skill transfer and simulation. In consumer technology, it drives AR, gaming, and social interaction. While multi-camera setups offer precision, they are impractical at scale. Monocular methods offer portability and accessibility, but only if they achieve sufficient accuracy. This makes the problem both scientifically important and practically impactful.

\begin{figure}
    \includegraphics[width=0.32\columnwidth]{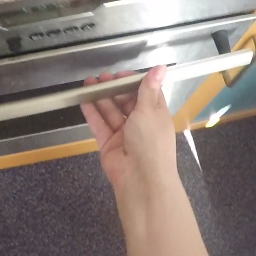}
    \begin{tikzpicture}
    \node[anchor=south west, inner sep=0] (image) at (0,0)
      {\includegraphics[width=0.32\columnwidth]{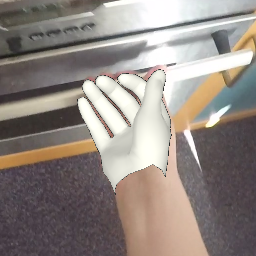}};

    \begin{scope}[x={(image.south east)}, y={(image.north west)}]
    \node[anchor=south west, inner sep=0pt,draw=red,line width=1mm] at (0.25, 0.8) {
    \includegraphics[width=0.07\textwidth,
                        trim=75 140 90 60, clip]{fig/res/hamer/TEST_epick_img_EK_0078_P02_09_frame_0000038575_l} 
      };
\end{scope}
\end{tikzpicture}
\begin{tikzpicture}   
\node[anchor=south west, inner sep=0] (image) at (0,0)
      {\includegraphics[width=0.32\columnwidth]{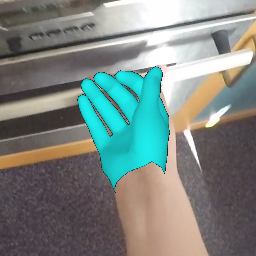}};

\begin{scope}[x={(image.south east)}, y={(image.north west)}]
\node[anchor=south west, inner sep=0pt,draw=green,line width=1mm] at (0.25, 0.8) {
\includegraphics[width=0.07\textwidth,
                         trim=75 140 90 60, clip]{fig/res/ours/TEST_epick_img_EK_0078_P02_09_frame_0000038575_l} 
      };
\end{scope}
\end{tikzpicture}

 \caption{What is otherwise a 3D estimation of high quality, clearly has room for improvement, as indicated by the imperfect match of the prediction and the observation (middle, estimated by HaMeR~\cite{pavlakos2024reconstructing}). This can be remedied (right, our method).}
 \label{fig:observation}
\vspace{-.5cm}
\end{figure}

Our starting point is a simple observation. Even high-performing models often produce hand reconstructions that do not perfectly align with the observed image appearance (see \cref{fig:observation}). This discrepancy suggests that current training signals, while effective, do not fully exploit available information. Motivated by this observation, we examine whether adding explicit, spatially grounded supervision based on image appearance can improve accuracy. Instead of treating texture as a byproduct, we use it as a source of alignment feedback. By linking pixels to predicted surfaces, we provide the model with a direct mechanism to correct its estimates.

With the main driving observation being misalignment, we examine a route towards dense alignment at training time. We approach this with a pixel-wise alignment loss between the observed appearance and the predicted hand surface. Computing such a loss requires three components: a 3D hypothesis of the hand (provided by an image-to-3D method), a model of its texture (the focus of our work), and a rendering mechanism (readily available through differentiable rendering libraries, e.g. \cite{ravi2020pytorch3d}) to project the textured mesh back onto the image. This setup allows us to transform discrepancies between prediction and observation into a usable training signal, and quantify its impact.

Dense, high-quality hand texture models are not readily available. Existing annotated datasets tend to provide full texture supervision only under multiview or studio conditions, which do not scale and often lack appearance diversity. In contrast, unannotated monocular data is abundant but provides only partial and imperfect observations. To bridge this gap, we construct texture supervision directly from monocular images using predictions from a pretrained 3D hand model. By projecting visible pixels onto the mesh surface and collecting them in UV space, we generate partial, view-specific texture maps without requiring ground truth UVs. These partial maps are then embedded into a learnable texture representation (see \cref{fig:teaser}). To that end, we employ a transformer architecture that attends directly to sparse UV-RGB observations. This module accepts an arbitrary number of input pixels and produces a complete, embedded texture. The flexibility of this design allows it to operate across varying levels of visibility and density, making it well-suited for learning from in-the-wild images. This approach eliminates the need for dense manual annotation, supports training from uncurated monocular data, and enables scalable learning from partial signals. We integrate this module into an existing monocular reconstruction pipeline by attaching it to a model that maps an input image to a 3D hand mesh (see \cref{fig:texturemodule}). During training, the mesh is projected back onto the image, and the visible surface points are used to extract UV-RGB pairs. Although the initial 3D prediction may contain errors, the transformer embeds the resulting sparse observations into a complete texture representation. This enables the model to generate coherent textures from incomplete inputs, handling diverse visibilities and observation densities typical of in-the-wild data. 

\vspace{1ex}
\noindent \textbf{Our main contributions are:}
\begin{itemize}
    \item We introduce the first method to accommodate sparse, partial hand texture observations into a unified model for full texture reconstruction, trained via weak-supervision and without requiring ground-truth textures or multiview input.

    \item We design a transformer-based architecture with pixel-level attention that reconstructs dense, coherent textures from incomplete UV-RGB inputs. To facilitate training from real-world data, we propose a scalable warm-up procedure that initializes the model using a mix of synthetic and real-world data, without requiring manual annotations.

    \item We employ a differentiable rendering framework and photometric consistency losses to supervise texture synthesis end-to-end, eliminating the need for manual annotations.

    \item We integrate our texture-guided supervision into a state-of-the-art 3D hand reconstruction pipeline, achieving a measurable improvement.
\end{itemize}

%% file: sec/2_related.tex
\section{Related Work}
\paragraph{Hand pose and shape estimation:}
A wide range of methods have been proposed to estimate hand pose and shape from visual data, often relying on parametric models or voxel-based representations. Approaches such as HAMBA~\cite{dong2024hamba}, Mesh Graphormer~\cite{lin2021mesh} and HandOS~\cite{chen2025handos} leverage transformer architectures and mesh-based representations to recover detailed hand shape and pose. Self-supervised and weakly-supervised learning strategies have also been explored to reduce reliance on annotated data~\cite{spurr2021self,moon2023bringing}. Earlier methods like~\cite{ge20193d,avola20223d,boukhayma20193d} focus on regressing 3D hand parameters or directly estimating mesh vertices from single images. Voxel-based techniques, such as HandVoxNet and its improved variant HandVoxNet++~\cite{malik2020handvoxnet,malik2021handvoxnet++}, utilize volumetric occupancy grids to represent hand geometry in 3D space. Collectively, these methods demonstrate ongoing progress in balancing accuracy, generalizability, and data efficiency in hand pose and shape estimation.
\paragraph{Hand texture models:}
Modeling realistic hand textures is essential for photorealistic rendering and accurate hand appearance reconstruction. HTML~\cite{qian2020html} introduces a generative framework that synthesizes detailed hand textures from monocular images, enabling high-fidelity appearance modeling. Handy~\cite{potamias2023handy} presents a neural parametric hand model that jointly captures geometry and texture in a coherent manner, enhancing realism across diverse viewing conditions. NIMBLE~\cite{li2022nimble} further extends this line of work by integrating a unified model of hand shape, pose, and appearance, offering controllable and expressive hand synthesis. These approaches collectively contribute to the development of comprehensive models that go beyond geometry to faithfully represent hand appearance in diverse scenarios.
\paragraph{Hand reconstruction:}
Methods based on convolutional neural networks~\cite{zhu2023hifihr,kim2024bitt} build on top of existing texture models to incorporate photometric loss. Chen et al.~\cite{chen2021model} propose to regress hand texture via MLP layers by taking a global feature vector as input. However, this approach handles occlusions poorly. Recent approaches have explored generative and reconstruction techniques for modeling human hands in 3D. Diffusion-based models~\cite{cheng2024handdiff, narasimhaswamy2024handiffuser}, have shown promising results in synthesizing realistic hand poses and appearances through probabilistic generative pipelines. Neural Radiance Fields (NeRFs)~\cite{mildenhall2021nerf} have also been adopted for hand modeling, offering view-consistent and detailed reconstructions~\cite{corona2022lisa,zhou2023nerf,guo2023handnerf,choi2024handnerf}. These methods leverage volumetric rendering and neural representations to capture complex hand geometries and articulation. However, despite their effectiveness, many of these approaches remain computationally expensive and resource-intensive, limiting their practicality in real-time or resource-constrained environments~\cite{Karvounas2021,karvounas2023dynamic,chen2024urhand}. This motivates the need for more efficient yet accurate hand modeling techniques.
\paragraph{Hand datasets:}
Researchers use a wide variety of datasets to train hand pose methods. These datasets provide 3D annotations, captured in both indoor and outdoor settings. Featuring a multi-camera set up, InterHand2.6M~\cite{moon2020interhand2} focuses on the strong interaction of hands. FreiHAND~\cite{zimmermann2019freihand}, HO-3D~\cite{hampali2020honnotate} and DexYCB~\cite{chao2021dexycb} are similarly captured under multi-camera studio settings, with emphasis on hand-object interactions. It is common to augment the training process using also datasets captured in the Panoptic Studio~\cite{simon2017hand, xiang2019monocular, joo2015panoptic}. Additionally, AssemblyHands~\cite{ohkawa2023assemblyhands} provides 3D ground truth while participants perform various assembly and disassembly tasks, again in a multi-camera setting. Despite the importance of 3D annotation, 2D annotation still plays a critical role. Therefore, large scale datasets like COCO-WholeBody~\cite{jin2020whole} and Halpe~\cite{fang2022alphapose} are commonly used~\cite{pavlakos2024reconstructing, chen2025handos,yang2023effective}. HInt~\cite{pavlakos2024reconstructing} was recently introduced, containing 2D annotated images from 3 egocentric datasets, NEWDAYS~\cite{cheng2023towards}, VISOR~\cite{darkhalil2022epic} and Ego4D~\cite{grauman2022ego4d}.


%% file: sec/3_method.tex
\section{Method}


Our goal is to improve monocular 3D hand reconstruction by integrating appearance-based supervision into an existing geometry-driven pipeline. To this end, we augment a baseline model that predicts 3D hand meshes from images (referred to as \textbf{BaseNet}) with a texture module that reconstructs UV-space textures from sparse, view-specific observations. This enables us to render a textured hand back into the input image and define a dense photometric loss that guides geometric refinement. The method consists of three components: the base reconstruction model, the texture model, and the associated training procedures. We describe each in turn.
\begin{figure*}[!htbp]
  \centering
  \includegraphics[clip, trim=16cm 20cm 28cm 13cm,width=\linewidth]{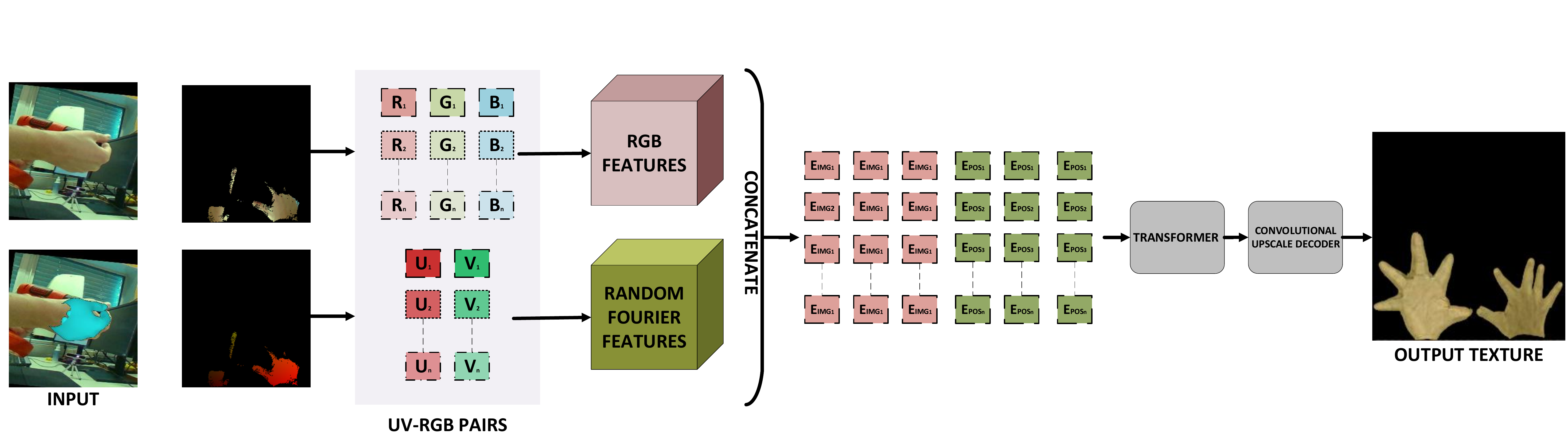}
  \caption{
  Our texture model consolidates variable-length, sparse, view-specific UV-RGB observations, extracted from a predicted 3D hand mesh and camera (BaseNet), into a complete hand texture. RGB values are embedded via a convolutional layer, and UV coordinates are encoded using Random Fourier Features~\cite{tancik2020fourier}. These feature pairs are concatenated and processed by a transformer decoder. A convolutional upscaling decoder synthesizes the final, dense texture map.
  } 
  \label{fig:texturemodule}
  \vspace{-.5cm}
\end{figure*}

\subsection{Hand Texture Model}

The texture model transforms sparse and potentially noisy appearance observations into a coherent UV-space representation, enabling dense, image-space supervision that improves the alignment and accuracy of the predicted hand geometry.

The texture model operates on a set of UV-space appearance observations 
$\mathcal{S} = \{ (u_i, c_i) \}_{i=1}^L$, where each $u_i \in [0,1]^2$ 
denotes a UV coordinate and $c_i \in \mathbb{R}^3$ is the corresponding RGB color 
sampled from the input image. Given this variable-length input, the model predicts 
a dense UV texture map $\hat{\mathcal{T}} \in \mathbb{R}^{3 \times H_T \times W_T}$, 
aligned with the fixed UV layout of the hand mesh.

In terms of architecture, we are faced with the following challenge: we are required to operate on sparse, pixel-perfect, variable-length input, and on a structured, image-based output. To that end, we employ a transformer-based encoder that attends to pixels, coupled with a convolutional upscaler, to yield the final full hand texture.

\paragraph{Transformer-Based Encoder:}

Given the sparse observations $\mathcal{S}$, we embed the RGB information via a $1 \times 1$ convolutional layer to produce an initial feature map. This is reshaped into a sequence of $L$ tokens of dimension $D$:
\begin{equation}
\mathbf{E}_{\text{img}} = \text{Conv1D}(S) \in \mathbb{R}^{ L \times D}.
\label{eq:mentionsD}
\end{equation}

To inject spatial context, we use a random Fourier encoding~\cite{tancik2020fourier} $\gamma(\cdot)$ applied to UV coordinates, that corresponds to each pixel, inputs $C \in \mathbb{R}^{ 2 \times L}$, yielding positional encodings $\mathbf{E}_{\text{pos}} \in \mathbb{R}^{ L \times D'}$:
\begin{equation}
\mathbf{E}_{\text{pos}} = \gamma(C).
\end{equation}
We concatenate these encodings:
\begin{equation}
\mathbf{E} = [\mathbf{E}_{\text{img}} \, \| \, \mathbf{E}_{\text{pos}}] \in \mathbb{R}^{ L \times (D + D')}.
\end{equation}

To aggregate global features, we introduce a set of $T$ learned query tokens $Q \in \mathbb{R}^{T \times (D + D')}$ that attend to $\mathbf{E}$ using a transformer. This is implemented using the TransformerDecoder implementation from PyTorch with $\mathbf{K}$ layers:
\begin{equation}
\mathbf{Z} = \text{Transformer}(Q, \mathbf{E}^\top),
\label{eq:mentionsZ}
\end{equation}
where $\mathbf{Z} \in \mathbb{R}^{ T \times D''}$ contains the encoded scene information and $\mathbf{E}$ serves as key and value.

\paragraph{Convolutional Upscaling Decoder:}

We employ a decoder to convert the latent representation \(\mathbf{Z}\)  into a full-resolution texture. It progressively upsamples the features using pixel shuffling operations~\cite{shi2016real}, which efficiently refine image details while increasing the spatial resolution, ultimately reaching \(224 \times 224\) pixels.

Each stage refines the feature representation, and the final convolution projects the upscaled features into RGB space, yielding the final hand texture $\hat{T} \in \mathbb{R}^{3 \times 224 \times 224}$ using 5 upscaling layers.

\paragraph{Weakly supervised texture learning:}

As explained in \cref{sec:results}, the most beneficial fine-tuning strategy involves pre-training the texture model and using it in a fixed inference mode during the subsequent fine-tuning of BaseNet.

We train the texture model in a weakly supervised setting using partial appearance observations extracted from monocular images. For each training image $I$, we obtain a predicted mesh and camera $(\hat{M}, \hat{c}) = f_{\text{base}}(I)$ and extract a set of UV-RGB samples $\mathcal{S} = \{ (u_i, c_i) \}_{i=1}^L$ by projecting visible mesh vertices into the image and associating their appearance with UV coordinates. These samples define supervision only over the visible subset of the UV domain. The texture model is trained to reconstruct a full texture $\hat{\mathcal{T}} = T(\mathcal{S})$ that agrees with the observed colors at their corresponding UV positions.

We define a sparse supervision map $T^*$ in UV space by assigning each $u_i$ to its observed color $c_i$, and a binary mask $\mathcal{M} \in \{0,1\}^{H_T \times W_T}$ indicating which texels are observed. The loss consists of a masked $L_1$ term and a Fourier-domain consistency loss:
\begin{align}
\mathcal{L}_{\text{weak}} &= 
\left\| \mathcal{M} \cdot (\hat{\mathcal{T}} - T^*) \right\|_1 +
\lambda_{\text{freq}} \cdot \mathcal{L}_{\text{Fourier}}, \\
\mathcal{L}_{\text{Fourier}} &= 
\left\| |\mathcal{F}(\hat{\mathcal{T}} \cdot \mathcal{M})| - |\mathcal{F}(T^* \cdot \mathcal{M})| \right\|_1,
\end{align}
where $\mathcal{F}(\cdot)$ denotes the discrete Fourier transform applied independently to each channel. This combination encourages both local accuracy in observed regions and global consistency in frequency space, helping the model produce coherent textures even from sparse and incomplete inputs.

The training samples are pseudo-labeled. In part, random samples are drawn directly on the texture space of parametric texture models, \eg, the HTML model \cite{qian2020html} in our case. The majority amount to partial textures extracted by back-projecting poses inferred by BaseNet, on real data, including the imperfections.

During training, we randomize the overall color of the texture, to encourage the model not to embed too far away from corner cases, \eg tattoos, gloves, occluders, \etc.

\subsection{Extending monocular 3D reconstruction}

We consider a monocular hand reconstruction system, denoted \textbf{BaseNet}, that maps an input image $I \in \mathbb{R}^{3 \times H \times W}$ to a 3D hand mesh and camera estimate, written as $(\hat{M}, \hat{c}) = f_{\text{base}}(I)$. We assume that the predicted mesh $\hat{M}$ is defined over a fixed topology with a consistent UV mapping. Using the image, predicted geometry, and camera, we extract UV-RGB correspondences $\mathcal{S} = \mathcal{P}(I, \hat{M}, \hat{c})$ by back-projecting visible surface points into the image. These sparse samples are passed to a texture module, which predicts a dense UV-space texture $\hat{\mathcal{T}} = T(\mathcal{S})$. The textured mesh is rendered using a differentiable renderer, yielding a synthetic image $\hat{I}_{\text{hand}} = \mathcal{R}(\hat{M}, \hat{\mathcal{T}}, \hat{c})$ that is compared to the input appearance. For explanatory clarity, we present the entire supervision pipeline in a simplified and compositional form. The resulting training objective is:
\begin{equation}
\mathcal{L}_{\text{total}} = 
\underbrace{\mathcal{L}_{\text{base}}}_{\text{BaseNet}} + 
\lambda_{\text{tex}} \cdot 
\underbrace{\mathcal{L}_{\text{tex}}(I, \hat{I}_{\text{hand}})}_{\text{Texture path}}.
\end{equation}

In this work, we instantiate \textbf{BaseNet} using HaMeR~\cite{pavlakos2024reconstructing}, a high-performing unadorned transformer-based model for monocular 3D hand reconstruction. HaMeR predicts hand pose, shape, and camera parameters, regressing the MANO~\cite{MANO:SIGGRAPHASIA:2017} model from a single image. For differentiable rendering, we use PyTorch3D~\cite{ravi2020pytorch3d}, which provides efficient projection and rasterization of textured meshes. Overall, to get an improved method, we fine-tune BaseNet by also incorporating 
the aforementioned changes.

\subsection{Putting it all together}
With our texture model as a sidecar to BaseNet, we perform a fine-tuning task. We re-train a pre-trained BaseNet, by maintaining its backbone fixed and only revisiting the head. Unless otherwise specified (see ablations in \cref{sec:results}), our texture model remains fixed, as well.

To the original BaseNet losses we simply amend our own. Otherwise, in terms of configuration and datasets, we follow the original BaseNet training strategy. During training, BaseNet learns to adjust its head so as to incur less alignment error on the training data. Finally, the texture module consists of 1.1 million parameters, with each training pass taking approximately 70 milliseconds.



\begin{figure}[htbp]
   \centering
   \includegraphics[clip, trim=22cm 13cm 2cm 7cm,width=\linewidth]{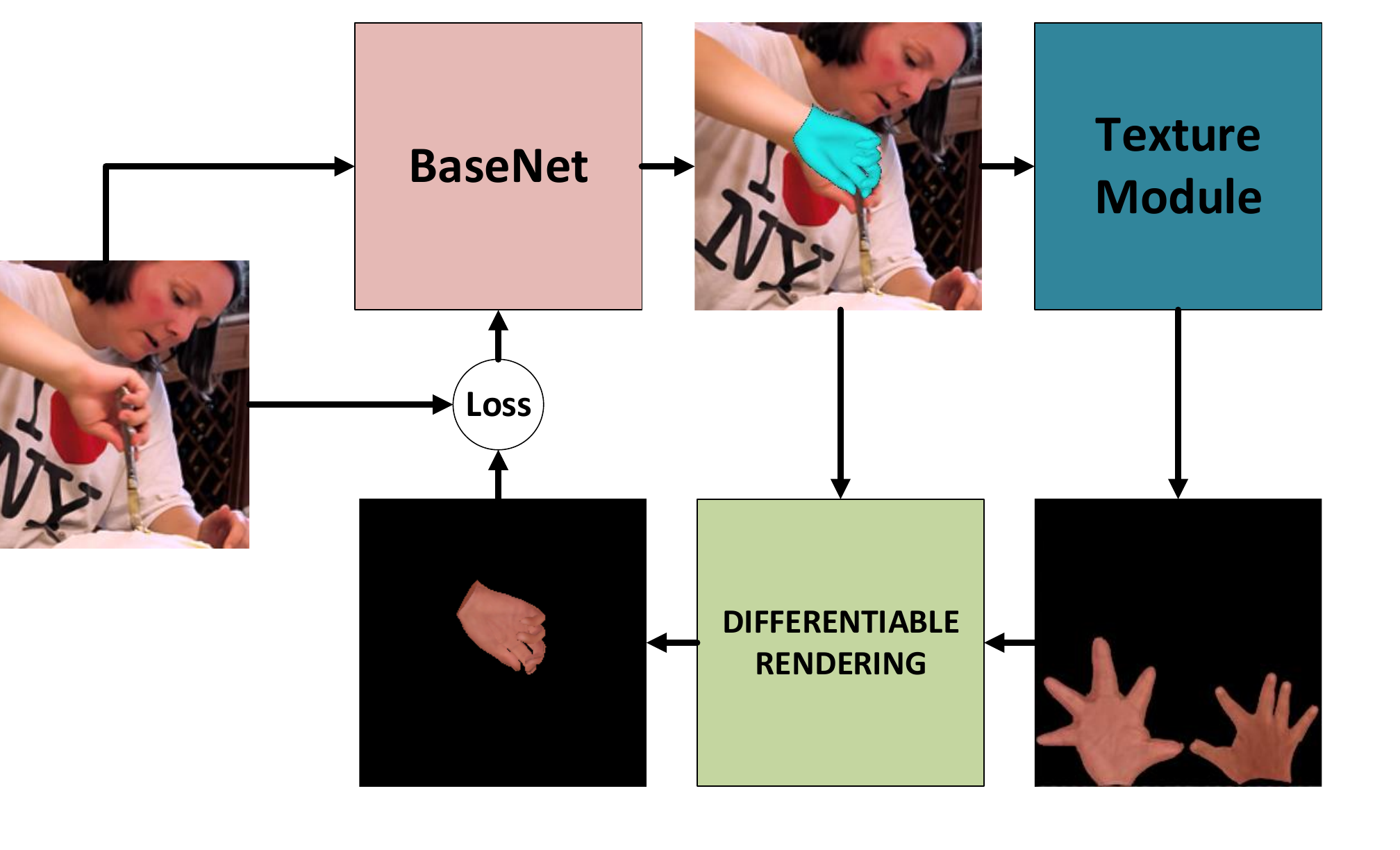}
   \caption{
   During training, the current 3D estimate from BaseNet is used to extract a noisy partial hand texture from the observation. This is refined through our texture model and is used to render a textured 3D hypothesis. The difference between the observation and the rendering comprises our additional loss term.
   }
   \label{fig:hamerntexture}
\vspace{-.5cm}
\end{figure}

%% file: sec/4_results.tex
\section{Results}
\label{sec:results}
\begin{table*}[!htbp]
\centering
\small
\setlength{\tabcolsep}{6pt}
\renewcommand{\arraystretch}{1.2}

\label{tab:performance_comparison}
\begin{tabular}{llccc ccc ccc}
\toprule
\multirow{2}{*}{Category} & \multirow{2}{*}{Method} 
  & \multicolumn{3}{c}{NEWDAYS} & \multicolumn{3}{c}{VISOR} & \multicolumn{3}{c}{Ego4D} \\
  \cmidrule(lr){3-5} \cmidrule(lr){6-8} \cmidrule(lr){9-11}
  & & @0.05 & @0.10 & @0.15 & @0.05 & @0.10 & @0.15 & @0.05 & @0.10 & @0.15 \\
\midrule

\multirow{3}{*}{All} 
&HaMeR~\cite{pavlakos2024reconstructing} &49.4 &79.3 &89.8 &44.4 &77.5 &89.7 &40.3 &72.4 &85.2 \\
&Ours &\bf49.5 &\bf79.5 &\bf89.9 &\bf46.8 &\bf79.1 &\bf90.6 &\bf42.6 &\bf73.5 &\bf86.1 \\

\midrule

\multirow{3}{*}{Visible} 
&HaMeR~\cite{pavlakos2024reconstructing} &\bf62.2 &\bf89.0 &\bf95.1 &58.5 &88.4 & 95.0 & 53.9 &84.2 &91.8 \\
&Ours &61.9 &88.8  & 95.0  &\bf61.2 &\bf89.0 &\bf95.3 &\bf57.2 &\bf84.8 &\bf92.3 \\

\midrule

\multirow{3}{*}{Occluded} 
  & HaMeR~\cite{pavlakos2024reconstructing} &28.4 &62.4 &80.1 &26.9 &61.8 &81.2 &24.3 &58.7 &77.3\\
  &Ours &\bf29.3 &\bf63.0 &\bf80.4 &\bf29.4 &\bf64.2 &\bf82.8 &\bf26.2 &\bf60.1 &\bf78.8 \\

\bottomrule
\end{tabular}
\caption{Comparison between HaMeR and our method on the Hint dataset at different thresholds. The proposed approach achieves better performance on this dataset. While HaMeR does outperform the proposed approach on the ``visible keypoints'' of the NEWDAYS subset, the difference is marginal.}
\label{tab:hamervsours}
\vspace{-0.5cm}
\end{table*}

We evaluate our approach in two parts. First, we demonstrate the positive effect of incorporating our appearance-based supervision into an existing monocular 3D hand reconstruction pipeline. Then, we take a closer look at certain aspects that are rather intrinsic to the texture model itself. 

Throughout, we adopt a controlled setup, keeping the backbone frozen and using a lightweight texture module to focus the analysis on the added value of texture-guided signals. The results show consistent improvements across datasets and metrics, suggesting that, even under conservative conditions, appearance supervision offers a meaningful complement to geometry-driven training, establishing the foundation on which more elaborate improvements can be founded (see \cref{sec:conclusion}).

 Experiments were conducted on a dual-socket AMD EPYC 7513 server with 64 physical cores, 1~TB RAM, and 4$\times$ NVIDIA A100-SXM4-80GB GPUs using CUDA 12.4. For this work we only employed one of the GPUs.

\subsection{Contrastive evaluation}
\label{secl:contrast}

We instantiate BaseNet with HaMeR~\cite{pavlakos2024reconstructing}, and throughout this work, the two terms are used interchangeably. For evaluation, we follow the experimental setup of HaMeR, utilizing the HInt dataset, which consists of 2D-annotated images drawn from three egocentric datasets: NEWDAYS~\cite{cheng2023towards}, VISOR~\cite{damen2018scaling}, and Ego4D~\cite{grauman2022ego4d}. Performance is assessed using the PCK metric~\cite{yang2012articulated}, consistent with HaMeR’s protocol.

To ensure a fair comparison and to attribute observed effects solely to our texture module, we deliberately refrain from retraining the HaMeR backbone, which comprises over 600M parameters. Given such high capacity, retraining could overshadow the impact of our comparatively lightweight module, making it difficult to isolate the contribution of texture-based supervision.

\begin{table}[htbp]
\resizebox{\columnwidth}{!}{%
\begin{tabular}{@{}lcccc@{}}
\toprule
\centering
Method &  PA- &  PA- &  F@5 $\uparrow$ & F@15 $\uparrow$ \\
 & MPJPE $\downarrow$ & MPVPE $\downarrow$ &  &  \\ \midrule
I2L-MeshNet~\cite{moon2020i2l} & 7.4 & 7.6 & 0.681 & 0.973 \\
Pose2Mesh~\cite{choi2020pose2mesh} & 7.7 & 7.8 & 0.674 & 0.969 \\
I2UV-HandNet~\cite{chen2021i2uv} & 6.7 & 6.9 & 0.707 & 0.977 \\
METRO~\cite{lin2021end} & 6.5 & 6.3 & 0.731 & 0.984 \\
Tang~\etal~\cite{tang2021towards} & 6.7 & 6.7 & 0.724 & 0.981 \\
Mesh Graphormer~\cite{lin2021mesh} & 5.9 & 6.0 & 0.764 & 0.986 \\ 
MobRecon~\cite{chen2022mobrecon} & {\bf 5.7} & 5.8 & 0.784 & 0.986 \\
AMVUR~\cite{jiang2023probabilistic} & 6.2 & 6.1 & 0.767 & 0.987 \\ \midrule
HaMeR~\cite{pavlakos2024reconstructing} & 6.0 & \bf 5.7 & {\bf 0.785} & {\bf 0.990} \\ 
Ours & 6.1 &  {\bf5.7} & {\bf 0.785} & {\bf 0.990} \\ \bottomrule
\end{tabular}
}
\caption{The proposed method achieves performance comparable to that of the initial method on the FreiHAND~\cite{zimmermann2019freihand} dataset.}
\label{tab:freihand}
\vspace{-.5cm}
\end{table}

\subsubsection{Quantitative analysis}
\label{sec:quantitative}

To quantify the effect of our texture model, as used through our dense alignment loss, we evaluated the HaMeR validation protocol~\cite{pavlakos2024reconstructing} on the original implementation of HaMeR and our training-time enhancement. The findings in \cref{tab:hamervsours} indicate that our approach consistently outperforms HaMeR in the ``All'' and ``Occluded'' keypoint categories across all datasets and thresholds, with the most pronounced improvements observed for occluded keypoints. For instance, on VISOR at @0.10, our method achieves 64.2\% PCK for occluded keypoints, compared to 61.8\% for HaMeR. This trend is consistent across all occluded cases, demonstrating the effectiveness of our texture-guided supervision in scenarios with limited visibility. In the ``Visible'' category, both methods perform comparably: HaMeR obtains slightly higher scores on the NEWDAYS subset (maximum difference of 0.3), while our method attains the best or equal results on VISOR and Ego4D. Overall, these results validate the benefit of incorporating appearance-based alignment, particularly in the presence of occlusions, while maintaining competitive performance in visible regions.

\begin{table}[htbp]
\resizebox{\columnwidth}{!}{
  \centering
  \scriptsize
\begin{tabular}{@{}lcccccc@{}}
\toprule
Method & $\textrm{AUC}_\textrm{J}$ $\uparrow$ & PA- & $\textrm{AUC}_\textrm{V}$ $\uparrow$ & PA- & F@5 $\uparrow$ & F@15 $\uparrow$ \\ 
   &  & MPJPE $\downarrow$ &  & MPVPE $\downarrow$ &  &  \\ \midrule

Liu~\etal~\cite{liu2021semi} & 0.803 & 9.9 &0.810&9.5&0.528&0.956\\
HandOccNet~\cite{park2022handoccnet} & 0.819 & 9.1 &0.819&8.8&0.564&0.963\\
I2UV-HandNet~\cite{chen2021i2uv} & 0.804 & 9.9 &0.799&10.1&0.500&0.943\\
Hampali~\etal~\cite{hampali2020honnotate} & 0.788 & 10.7 &0.790&10.6&0.506&0.942\\
Hasson~\etal~\cite{hasson2019learning} & 0.780 & 11.0 &0.777&11.2&0.464&0.939\\
ArtiBoost~\cite{yang2022artiboost} & 0.773 & 11.4 &0.782&10.9&0.488&0.944\\
Pose2Mesh~\cite{choi2020pose2mesh} & 0.754 & 12.5 &0.749&12.7&0.441&0.909\\
I2L-MeshNet~\cite{moon2020i2l} & 0.775 & 11.2 &0.722&13.9&0.409&0.932\\
METRO~\cite{lin2021end} & 0.792 & 10.4 & 0.779 & 11.1 & 0.484 & 0.946\\
MobRecon\cite{chen2022mobrecon}& -&9.2&-& 9.4& 0.538& 0.957\\
Keypoint Trans~\cite{hampali2022keypoint} & 0.786 & 10.8 &-&-&-&-\\
AMVUR~\cite{jiang2023probabilistic} & 0.835 & 8.3 & 0.836 & 8.2 & 0.608 & 0.965 \\ \midrule
HaMeR & {\bf 0.846} & {\bf 7.7} & {\bf 0.841} & {\bf 7.9} & {\bf 0.635} & {\bf 0.980} \\ 
Ours & {0.844} & {7.8} & {0.840} & {8.0} & {0.630} & {\bf0.980} \\ \bottomrule
\end{tabular}
}
    \caption{\textbf{Comparison with the state-of-the-art on the HO3D dataset~\cite{hampali2020honnotate}.} We use the HO3Dv2 protocol and report metrics that evaluate accuracy of the estimated
    3D joints and 3D mesh. PA-MPVPE and PA-MPJPE numbers are in mm.}
  \label{tab:ho3d}%
\vspace{-.25cm}
\end{table}%
\addtolength{\tabcolsep}{5pt}

Additionally, we include comparisons in standardized 3D benchmarks, such as the FreiHAND~\cite{zimmermann2019freihand} benchmark (see \cref{tab:freihand}) and the HO3Dv2~\cite{hampali2020honnotate,hampali2022keypointtransformer} benchmark (see \cref{tab:ho3d}). These results show a comparative performance that is on par with HaMeR, showing no regression on studio-captured, well-annotated, and relatively clean datasets, while at the same time improvement is achieved on egocentric, in-the-wild, or heavily occluded settings.

Similarly, our method also performs on par with another recent state-of-the-art baseline, HAMBA~\cite{dong2024hamba} (see \cref{fig:hamba}), which builds on a modified Mamba~\cite{mamba,mamba2} architecture with additional components and training strategies. Interestingly, we attain similar performance through a straightforward training-time addition of supervision on an otherwise unadorned transformer-based architecture.

\begin{table}[htbp]
\centering
\scriptsize
\setlength{\tabcolsep}{2.5pt}  
\renewcommand{\arraystretch}{1.15}
\label{tab:hint_comparison}
\begin{tabular}{
  >{\centering\arraybackslash}m{0.8em}
  l
  @{\,}c@{}c@{}c@{\quad}
  c@{}c@{}c@{\quad}
  c@{}c@{}c
}
\toprule
& \textbf{Method} 
  & \multicolumn{3}{c}{\textbf{New Days}} 
  & \multicolumn{3}{c}{\textbf{VISOR}} 
  & \multicolumn{3}{c}{\textbf{Ego4D}} \\
& 
  & @0.05 & @0.10 & @0.15 
  & @0.05 & @0.10 & @0.15 
  & @0.05 & @0.10 & @0.15 \\
\midrule

\multirow{2}{*}{\rotatebox[origin=c]{90}{\textcolor{green!50!black}{All}}} 
&HAMBA~\cite{dong2024hamba} &48.7 &79.2 &\bf90.0 &\bf47.2 &\bf80.2 &\bf91.2 &41.7 &72.9 &85.5 \\
&Ours &\bf49.5 &\bf79.5 &89.9 &46.8 &79.1 &90.6 &\bf42.6 &\bf73.5 &\bf86.1 \\

\midrule

\multirow{2}{*}{\rotatebox[origin=c]{90}{\textcolor{green!50!black}{Vis.}}} 
&HAMBA~\cite{dong2024hamba} &61.2 &88.4 &94.9 &\bf61.4 &\bf89.6 &\bf95.6 &56.0 &84.3 &91.9 \\
&Ours &\bf61.9 &\bf88.8  &\bf95.0  &61.2 &89.0 &95.3 &\bf57.2 &\bf84.8 &\bf92.3 \\

\midrule

\multirow{2}{*}{\rotatebox[origin=c]{90}{\textcolor{green!50!black}{Occl.}}} 
&HAMBA~\cite{dong2024hamba} &28.2 &62.8 &\bf81.1 &\bf29.9 &\bf66.6 &\bf84.3 &25.2  &59.2 &77.6\\
&Ours &\bf29.3 &\bf63.0 &80.4 &29.4 &64.2 &82.8 &\bf26.2 &\bf60.1 &\bf78.8\\
\bottomrule
\end{tabular}
\caption{Performance comparison between our method and HAMBA~\cite{dong2024hamba} on the HInt dataset. Best results are shown in \textbf{bold}.}
\vspace{-0.3cm}
\label{fig:hamba}

\end{table}

\begin{table}[t]
\centering
\scriptsize
\setlength{\tabcolsep}{2.5pt}  
\renewcommand{\arraystretch}{1.15}
\begin{tabular}{
  >{\centering\arraybackslash}m{0.8em}
  l
  @{\,}c@{}c@{}c@{\quad}
  c@{}c@{}c@{\quad}
  c@{}c@{}c
}
\toprule
& \textbf{Variant} 
  & \multicolumn{3}{c}{\textbf{New Days}} 
  & \multicolumn{3}{c}{\textbf{VISOR}} 
  & \multicolumn{3}{c}{\textbf{Ego4D}} \\
& 
  & @0.05 & @0.10 & @0.15 
  & @0.05 & @0.10 & @0.15 
  & @0.05 & @0.10 & @0.15 \\
\midrule

\multirow{3}{*}{\rotatebox[origin=c]{90}{\textcolor{green!50!black}{All}}}
  & H\&M*  &45.9 & 78.1 & 89.6 & 39.9 & 76.8 & 89.7 & 37.1 & 71.4 & 85.0 \\
  & H\&M &47.4 & 79.2 &\bf90.0 & 43.4 & 77.8 & 90.1 & 40.7 & 72.9 &85.6\\
  & H &\bf49.5 &\bf79.5 &89.9 &\bf46.8 &\bf79.1 &\bf90.6 &\bf42.6 &\bf73.5 &\bf86.1 \\
\midrule
\multirow{3}{*}{\rotatebox[origin=c]{90}{\textcolor{green!50!black}{Visible}}} 
  & H\&M* & 58.1 & 87.7 & 94.9 & 51.4 & 87.6 & 95.2 & 48.3 & 82.6 & 91.7\\
  & H\&M  & 58.7 & 88.5 &\bf95.1 & 56.5 & 88.4 &\bf 95.3 & 54.0 & 83.8 &92.1 \\
  & H     &\bf61.9 &\bf88.8  & 95.0  &\bf61.2 &\bf89.0 &\bf95.3 &\bf57.2 &\bf84.8 &\bf92.3 \\
\midrule
\multirow{3}{*}{\rotatebox[origin=c]{90}{\textcolor{green!50!black}{Occluded}}}
  & H\&M*   & 26.2 & 61.4 & 80.2 & 26.1 & 61.6 & 81.3 & 23.4 & 58.0 & 77.4\\
  & H\&M  & 29.0 & 62.7 & \bf81.0 & 27.6 & 62.8 & 81.6 & 25.7 &59.8 & 78.1\\ 
  & H      &\bf29.3 &\bf63.0 &80.4 &\bf29.4 &\bf64.2 &\bf82.8 &\bf26.2 &\bf60.1 &\bf78.8\\
\bottomrule
\end{tabular}
\caption{Performance comparison on the HInt dataset
with the head retrained using the texture module. We denote training the Head by H, training the head with the warmed up texture module by H\&M. Finally, H\&M* denotes training of the head and the texture module from scratch.}
\label{tab:hintcomparisonablation}
\vspace{-.3cm}
\end{table}

\subsubsection{Qualitative analysis}
\label{sec:qualitative}

\begin{figure*}[!ht]
    \includegraphics[width=0.32\columnwidth]{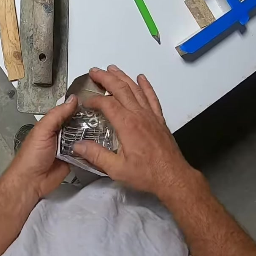}
    \begin{tikzpicture}
    \node[anchor=south west, inner sep=0] (image) at (0,0)
      {\includegraphics[width=0.32\columnwidth]{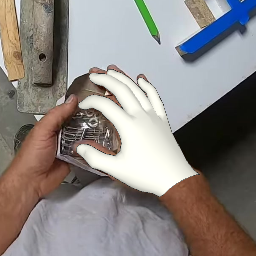}};

    \begin{scope}[x={(image.south east)}, y={(image.north west)}]
    \node[anchor=south west, inner sep=0pt,draw=red,line width=1mm] at (0.65, 0.6) {
    \includegraphics[width=0.05\textwidth,
                         trim=85 140 100 60, clip]{fig/res/hamer/TEST_ego4d_img_1b2f1101-b3d9-4be5-bc92-393c983be9b4_2771e3c7-0481-4953-b2d0-1f1ab951cf46_699_pre_15_r} 
      };
\end{scope}
\end{tikzpicture}
\begin{tikzpicture}   
\node[anchor=south west, inner sep=0] (image) at (0,0)
      {\includegraphics[width=0.32\columnwidth]{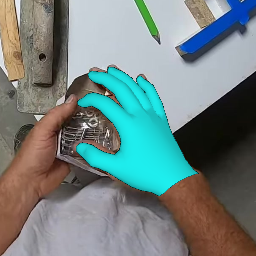}};

\begin{scope}[x={(image.south east)}, y={(image.north west)}]
\node[anchor=south west, inner sep=0pt,draw=green,line width=1mm] at (0.65, 0.6) {
\includegraphics[width=0.05\textwidth,
                         trim=85 140 100 60, clip]{fig/res/ours/TEST_ego4d_img_1b2f1101-b3d9-4be5-bc92-393c983be9b4_2771e3c7-0481-4953-b2d0-1f1ab951cf46_699_pre_15_r} 
      };
\end{scope}
\end{tikzpicture}
    \hfill  
    \includegraphics[width=0.32\columnwidth]{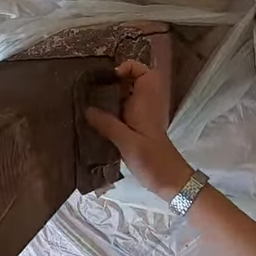}
    \begin{tikzpicture}
    \node[anchor=south west, inner sep=0] (image) at (0,0)
      {\includegraphics[width=0.32\columnwidth]{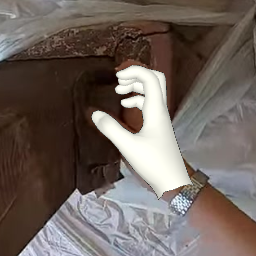}};

    \begin{scope}[x={(image.south east)}, y={(image.north west)}]
    \node[anchor=south west, inner sep=0pt,draw=red,line width=1mm] at (0.65, 0.6) {
    \includegraphics[width=0.05\textwidth,
                         trim=85 140 100 50, clip]{fig/res/hamer/TEST_ego4d_img_1d37b42e-2a9e-4ba9-aabc-7daac0c15118_275b6b68-a828-440f-a3c7-e208811ac158_2520_pre_frame_r} 
      };
\end{scope}
\end{tikzpicture}
\begin{tikzpicture}   
\node[anchor=south west, inner sep=0] (image) at (0,0)
      {\includegraphics[width=0.32\columnwidth]{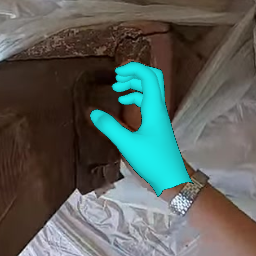}};

\begin{scope}[x={(image.south east)}, y={(image.north west)}]
\node[anchor=south west, inner sep=0pt,draw=green,line width=1mm] at (0.65, 0.6) {
\includegraphics[width=0.05\textwidth,
                         trim=85 140 100 50, clip]{fig/res/ours/TEST_ego4d_img_1d37b42e-2a9e-4ba9-aabc-7daac0c15118_275b6b68-a828-440f-a3c7-e208811ac158_2520_pre_frame_r} 
      };
\end{scope}
\end{tikzpicture}

\includegraphics[width=0.32\columnwidth]{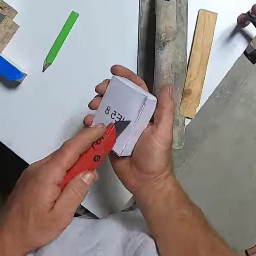}
\begin{tikzpicture}
    \node[anchor=south west, inner sep=0] (image) at (0,0)
      {\includegraphics[width=0.32\columnwidth]{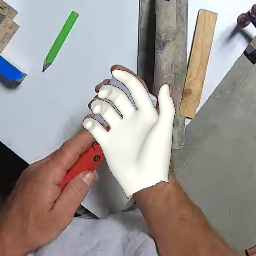}};

    \begin{scope}[x={(image.south east)}, y={(image.north west)}]
    \node[anchor=south west, inner sep=0pt,draw=red,line width=1mm] at (0.65, 0.6) {
    \includegraphics[width=0.05\textwidth,
                         trim=85 140 100 50, clip]{fig/res/hamer/TEST_ego4d_img_1b2f1101-b3d9-4be5-bc92-393c983be9b4_fc6b3573-ec47-4cdd-8bcf-457bbb4c3a2e_580_pre_30_l} 
      };
\end{scope}
\end{tikzpicture}
\begin{tikzpicture}   
\node[anchor=south west, inner sep=0] (image) at (0,0)
      {\includegraphics[width=0.32\columnwidth]{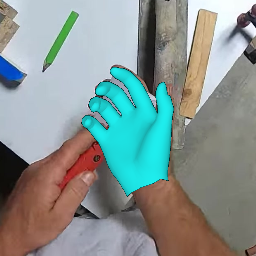}};

\begin{scope}[x={(image.south east)}, y={(image.north west)}]
\node[anchor=south west, inner sep=0pt,draw=green,line width=1mm] at (0.65, 0.6) {
\includegraphics[width=0.05\textwidth,
                         trim=85 140 100 50, clip]{fig/res/ours/TEST_ego4d_img_1b2f1101-b3d9-4be5-bc92-393c983be9b4_fc6b3573-ec47-4cdd-8bcf-457bbb4c3a2e_580_pre_30_l} 
      };
\end{scope}
\end{tikzpicture}
    \hfill
    \includegraphics[width=0.32\columnwidth]{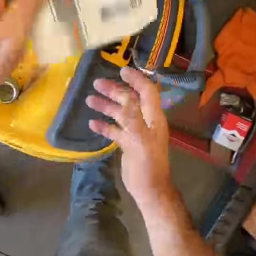}
\begin{tikzpicture}
    \node[anchor=south west, inner sep=0] (image) at (0,0)
      {\includegraphics[width=0.32\columnwidth]{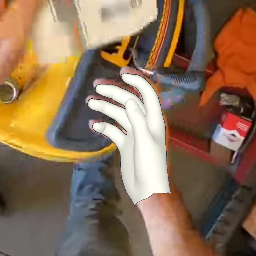}};

    \begin{scope}[x={(image.south east)}, y={(image.north west)}]
    \node[anchor=south west, inner sep=0pt,draw=red,line width=1mm] at (0.65, 0.6) {
    \includegraphics[width=0.05\textwidth,
                         trim=85 140 100 50, clip]{fig/res/hamer/TEST_ego4d_img_5a96bb31-3fff-4546-9fe8-d1343f2fe19f_b23341f4-824d-4999-8cc7-47141a3547b2_1104_pre_frame_r} 
      };
\end{scope}
\end{tikzpicture}
\begin{tikzpicture}   
\node[anchor=south west, inner sep=0] (image) at (0,0)
      {\includegraphics[width=0.32\columnwidth]{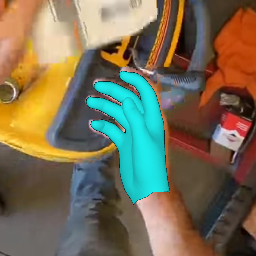}};

\begin{scope}[x={(image.south east)}, y={(image.north west)}]
\node[anchor=south west, inner sep=0pt,draw=green,line width=1mm] at (0.65, 0.6) {
\includegraphics[width=0.05\textwidth,
                         trim=85 140 100 50, clip]{fig/res/ours/TEST_ego4d_img_5a96bb31-3fff-4546-9fe8-d1343f2fe19f_b23341f4-824d-4999-8cc7-47141a3547b2_1104_pre_frame_r} 
      };
\end{scope}
\end{tikzpicture}

    \includegraphics[width=0.32\columnwidth]{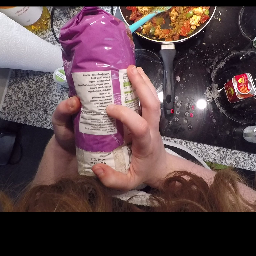}
\begin{tikzpicture}
    \node[anchor=south west, inner sep=0] (image) at (0,0)
      {\includegraphics[width=0.32\columnwidth]{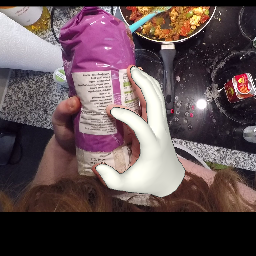}};

    \begin{scope}[x={(image.south east)}, y={(image.north west)}]
    \node[anchor=south west, inner sep=0pt,draw=red,line width=1mm] at (0.65, 0.6) {
    \includegraphics[width=0.05\textwidth,
                         trim=75 110 110 80, clip]{fig/res/hamer/TEST_epick_img_EK_0086_P04_13_frame_0000010091_l} 
      };
\end{scope}
\end{tikzpicture}
\begin{tikzpicture}   
\node[anchor=south west, inner sep=0] (image) at (0,0)
      {\includegraphics[width=0.32\columnwidth]{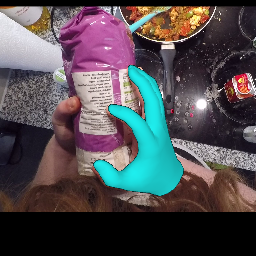}};

\begin{scope}[x={(image.south east)}, y={(image.north west)}]
\node[anchor=south west, inner sep=0pt,draw=green,line width=1mm] at (0.65, 0.6) {
\includegraphics[width=0.05\textwidth,
                         trim=75 110 110 80, clip]{fig/res/ours/TEST_epick_img_EK_0086_P04_13_frame_0000010091_l} 
      };
\end{scope}
\end{tikzpicture}
    \hfill
    \includegraphics[width=0.32\columnwidth]{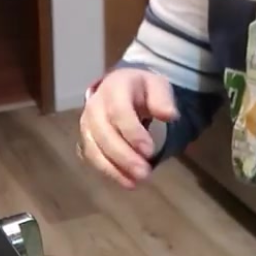}
\begin{tikzpicture}
    \node[anchor=south west, inner sep=0] (image) at (0,0)
      {\includegraphics[width=0.32\columnwidth]{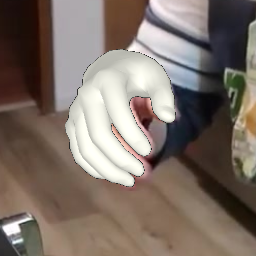}};

    \begin{scope}[x={(image.south east)}, y={(image.north west)}]
    \node[anchor=south west, inner sep=0pt,draw=red,line width=1mm] at (0.65, 0.6) {
    \includegraphics[width=0.05\textwidth,
                         trim=75 80 110 120, clip]{fig/res/hamer/TEST_newdays_img_ND_OdYBQcaXkO8_frame030001_l} 
      };
\end{scope}
\end{tikzpicture}
\begin{tikzpicture}   
\node[anchor=south west, inner sep=0] (image) at (0,0)
      {\includegraphics[width=0.32\columnwidth]{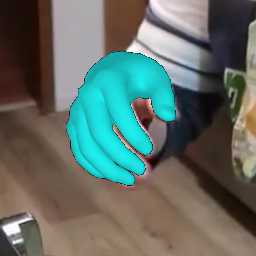}};

\begin{scope}[x={(image.south east)}, y={(image.north west)}]
\node[anchor=south west, inner sep=0pt,draw=green,line width=1mm] at (0.65, 0.6) {
\includegraphics[width=0.05\textwidth,
                         trim=75 80 110 120, clip]{fig/res/ours/TEST_newdays_img_ND_OdYBQcaXkO8_frame030001_l} 
      };
\end{scope}
\end{tikzpicture}

\includegraphics[width=0.32\columnwidth]{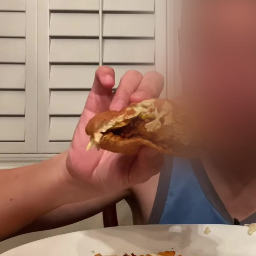}
\begin{tikzpicture}
    \node[anchor=south west, inner sep=0] (image) at (0,0)
      {\includegraphics[width=0.32\columnwidth]{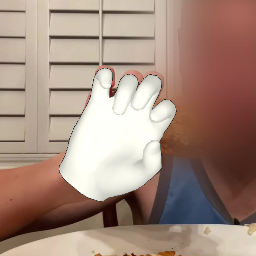}};

    \begin{scope}[x={(image.south east)}, y={(image.north west)}]
    \node[anchor=south west, inner sep=0pt,draw=red,line width=1mm] at (0.65, 0.6) {
    \includegraphics[width=0.05\textwidth,
                         trim=90 135 95 55, clip]{fig/res/hamer/TEST_newdays_img_ND_aTu8teDhHiw_frame023622_r} 
      };
\end{scope}
\end{tikzpicture}
\begin{tikzpicture}   
\node[anchor=south west, inner sep=0] (image) at (0,0)
      {\includegraphics[width=0.32\columnwidth]{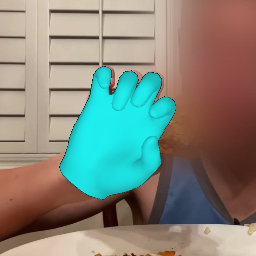}};

\begin{scope}[x={(image.south east)}, y={(image.north west)}]
\node[anchor=south west, inner sep=0pt,draw=green,line width=1mm] at (0.65, 0.6) {
\includegraphics[width=0.05\textwidth,
                         trim=90 135 95 55, clip]{fig/res/ours/TEST_newdays_img_ND_aTu8teDhHiw_frame023622_r} 
      };
\end{scope}
\end{tikzpicture}
    \hfill
    \includegraphics[width=0.32\columnwidth]{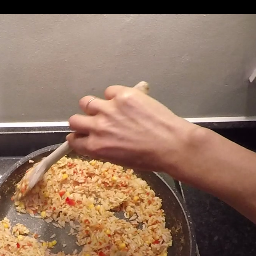}
\begin{tikzpicture}
    \node[anchor=south west, inner sep=0] (image) at (0,0)
      {\includegraphics[width=0.32\columnwidth]{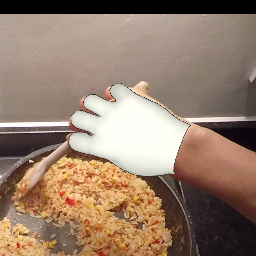}};

    \begin{scope}[x={(image.south east)}, y={(image.north west)}]
    \node[anchor=south west, inner sep=0pt,draw=red,line width=1mm] at (0.65, 0.6) {
    \includegraphics[width=0.05\textwidth,
                         trim=55 100 130 100, clip]{fig/res/hamer/TEST_epick_img_EK_0129_P08_17_frame_0000022734_r} 
      };
\end{scope}
\end{tikzpicture}
\begin{tikzpicture}   
\node[anchor=south west, inner sep=0] (image) at (0,0)
      {\includegraphics[width=0.32\columnwidth]{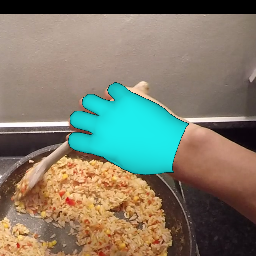}};

\begin{scope}[x={(image.south east)}, y={(image.north west)}]
\node[anchor=south west, inner sep=0pt,draw=green,line width=1mm] at (0.65, 0.6) {
\includegraphics[width=0.05\textwidth,
                         trim=55 100 130 100, clip]{fig/res/ours/TEST_epick_img_EK_0129_P08_17_frame_0000022734_r} 
      };
\end{scope}
\end{tikzpicture}

    \includegraphics[width=0.32\columnwidth]{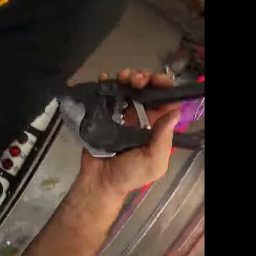}
\begin{tikzpicture}
    \node[anchor=south west, inner sep=0] (image) at (0,0)
      {\includegraphics[width=0.32\columnwidth]{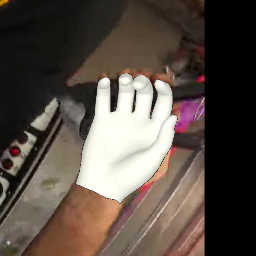}};

    \begin{scope}[x={(image.south east)}, y={(image.north west)}]
    \node[anchor=south west, inner sep=0pt,draw=red,line width=1mm] at (0.65, 0.6) {
    \includegraphics[width=0.05\textwidth,
                         trim=95 140 90 60, clip]{fig/res/hamer/TEST_ego4d_img_b527da99-9e7d-4109-b9eb-7ce483a620a3_1ca60f96-2aad-4886-9b08-9eeaf2c80f7e_5982_pre_30_l} 
      };
\end{scope}
\end{tikzpicture}
\begin{tikzpicture}   
\node[anchor=south west, inner sep=0] (image) at (0,0)
      {\includegraphics[width=0.32\columnwidth]{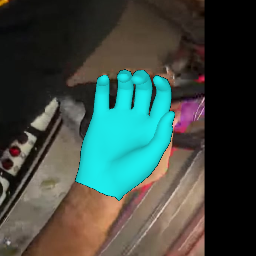}};

\begin{scope}[x={(image.south east)}, y={(image.north west)}]
\node[anchor=south west, inner sep=0pt,draw=green,line width=1mm] at (0.65, 0.6) {
\includegraphics[width=0.05\textwidth,
                         trim=95 140 90 60, clip]{fig/res/ours/TEST_ego4d_img_b527da99-9e7d-4109-b9eb-7ce483a620a3_1ca60f96-2aad-4886-9b08-9eeaf2c80f7e_5982_pre_30_l} 
      };
\end{scope}
\end{tikzpicture}
    \hfill
    \includegraphics[width=0.32\columnwidth]{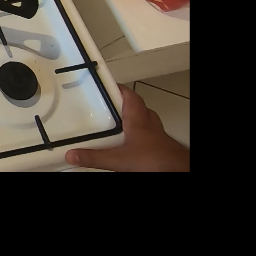}
\begin{tikzpicture}
    \node[anchor=south west, inner sep=0] (image) at (0,0)
      {\includegraphics[width=0.32\columnwidth]{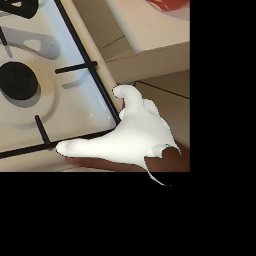}};

    \begin{scope}[x={(image.south east)}, y={(image.north west)}]
    \node[anchor=south west, inner sep=0pt,draw=red,line width=1mm] at (0.65, 0.6) {
    \includegraphics[width=0.05\textwidth,
                         trim=30 80 155 100, clip]{fig/res/hamer/TEST_epick_img_EK_0033_P25_101_frame_0000009798_l} 
      };
\end{scope}
\end{tikzpicture}
\begin{tikzpicture}   
\node[anchor=south west, inner sep=0] (image) at (0,0)
      {\includegraphics[width=0.32\columnwidth]{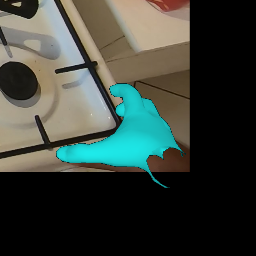}};

\begin{scope}[x={(image.south east)}, y={(image.north west)}]
\node[anchor=south west, inner sep=0pt,draw=green,line width=1mm] at (0.65, 0.6) {
\includegraphics[width=0.05\textwidth,
                         trim=30 80 155 100, clip]{fig/res/ours/TEST_epick_img_EK_0033_P25_101_frame_0000009798_l} 
      };
\end{scope}
\end{tikzpicture}

    \caption{
    Indicative qualitative results on the HInt dataset.
    The first column shows the input image. The second column presents the estimated hand prediction from HaMeR. The third column displays the predicted result from our method. Evidently the proposed method, using texture priors, can better estimate the hand shape and pose.}
    \label{fig:qualitative}
    \vspace{-.5cm}
\end{figure*}

From the results of the quantitative analysis (see \cref{sec:quantitative}) we draw samples that help understand what the quantitative improvement amounts to in terms of visual improvement and ``fit'' improvement. As shown in \cref{fig:qualitative},  dense alignment during training, enabled through our texture model, leads to ``fitter'' overlays of hypotheses over the corresponding observations. For example, a visually apparent improvement regards the better localization of the fingers.

\begin{table}[t]
\centering
\begin{tabular}{lcc}
\toprule
\textbf{\# Visible Pixels} & \textbf{L1} ↓ & \textbf{SSIM} ↑  \\
\midrule
200   & 7.6 & 0.85  \\
500   & 6.2  & 0.89 \\
1000  & 5.9  & 0.89 \\
2000  & 5.4 &  0.91 \\
ALL  & \textbf{3.3} & \textbf{0.95} \\
\bottomrule
\end{tabular}
\caption{Quantitative metrics on the effect the number of visible UV pixels has on texture reconstruction quality.}
\label{tab:pixel_count_ablation}
\vspace{-.5cm}
\end{table}

\subsection{Texture model insights}
\label{sec:insights}

We analyze how different configurations influence the performance of our method or the texture model itself. When it comes to texture quality or fidelity, we employ  the SSIM and L1 metrics and measure against known ground truth generated from HTML~\cite{qian2020html}.

\vspace{0.5em}
\noindent
\textbf{Initialize/Freeze?} We compare the following variants, \wrt fine-tuning HaMeR:
\begin{itemize}
    \item $H$: only fine-tune the head of HaMeR (the texture module is warmed up),
    \item $H\&M$: fine-tune both the head of HaMeR and the warmed up texture module, and,
    \item $H\&M*$: like $H\&M$, but with a randomly initialized texture model.
\end{itemize}
The results in \cref{tab:hintcomparisonablation} indicate that $H$ is the most effective approach in this context, and the one considered throughout. It is anticipated that with changes and/or in a different setting, the results might differ (see \cref{sec:conclusion}).

\vspace{0.5em}
\noindent
\textbf{Observation density VS ``recall'':}
Because each monocular observation offers only partial UV coverage, it is important to understand how the number of observed UV-pixel pairs affects the quality of texture reconstruction. To investigate this, we vary the number of visible UV samples provided to the texture module (see~\cref{tab:pixel_count_ablation}). We find that our method maintains high reconstruction fidelity with as few as 1{,}000 visible UV pixels. Both L1 loss and SSIM remain stable above this threshold, while reconstruction quality degrades more noticeably below it. This analysis shows that the proposed approach is robust to the sparse and incomplete supervision typical of in-the-wild monocular data, and remains practical in real-world conditions where full surface visibility is rare.

\vspace{0.5em}
\noindent
\textbf{Hyperparameter tuning:}
We search for the optimal architectural configuration for the texture module by varying the number of transformer layers $\mathbf{K}$ and the embedding dimension $D$ (see \cref{eq:mentionsD}). Table~\ref{tab:model_metrics} summarizes the impact of these choices, evaluated using L1 loss and SSIM. We find that SSIM remains consistently high (up to 0.95) across several configurations, indicating robust perceptual quality. However, L1 loss shows more variation, with the lowest value (3.35) achieved using 8 layers and an embedding dimension of 6. This suggests that increasing model depth and selecting moderate embedding sizes can yield incremental improvements in pixel-wise accuracy, while overall visual quality remains stable. This is the configuration used throughout.

\vspace{0.5em}
\noindent
\textbf{Navigating Towards Better Textures:}
Through hands-on experimentation, we quickly identified design choices that consistently yield sharper and more realistic textures. Using pixel shuffle for upsampling, incorporating Fourier-based positional encodings, and adopting a moderately deep transformer backbone emerged as particularly effective. Additionally, combining L1 and Fourier-domain losses proved crucial for producing high-fidelity textures. These choices allow the texture module to operate reliably even with limited supervision, making it suitable for single-view reconstruction tasks in practical scenarios.

\begin{table}[htb]
\centering
\begin{tabular}{|c|c|c|c|}
\hline
\textbf{\# of Layers} & \textbf{Emb. Dim} & \textbf{L1}↓ & \textbf{SSIM} ↑ \\
\hline
6 & 6 & 6.25 & 0.90 \\
8 & 8 & 5.45 & 0.90 \\
4 & 6 & 5.12 & 0.91 \\
4 & 8 & 4.36 & 0.94 \\
6 & 8 & 3.77 & 0.95 \\
8 & 6 & \bf3.35 & \bf0.95 \\

\hline
\end{tabular}
\caption{Performance metrics for different number of layers and embedding dimensions.}
\label{tab:model_metrics}
\vspace{-0.5cm}
\end{table}

%% file: sec/5_conclusion.tex
\section{Conclusions and future work}
\label{sec:conclusion}


We have shown that adding a lightweight, texture‑aware supervision module only during training delivers systematic improvements to monocular 3‑D hand reconstruction while leaving the runtime architecture unchanged—the only difference at test time is the improved weights. Injecting fine‑grained appearance cues into optimization increases PCK by up to $2.7\%$ on heavily occluded HInt frames and yields consistent, albeit smaller, gains on FreiHAND and HO‑3D. The effect is most pronounced in scenarios where geometry‑only cues are ambiguous—severe occlusion and egocentric viewpoints—underscoring the complementary value of texture information.

The proposed module is practical: it requires no extra manual annotation, reconstructs dense UV maps from as few as $1.000$ observed texels, adds $<70$ms per training iteration, and incurs zero test‑time cost. A moderate‑depth transformer with Fourier positional encodings and pixel‑shuffle up‑sampling strikes the best balance between accuracy and efficiency.

Our work provides a complete, modular solution for integrating learned texture priors into monocular 3D hand reconstruction pipelines. The results across multiple benchmarks confirm that the method achieves its core objectives: improving geometric and appearance accuracy in a weakly-supervised setting, without reliance on multi-view data or dense annotations. The framework is robust, scalable, and readily deployable in real-world scenarios.

Still, the tools and insights developed here naturally give rise to a number of promising extensions:

\begin{itemize}
    \item Applying the texture prior to a broader range of hand reconstruction backbones to further validate its generality.
    \item Developing more advanced differentiable rendering strategies to improve training stability and convergence, as advocated in \cite{Karvounas2021}.
    \item Leveraging the texture module as a denoiser for pose estimation errors through targeted weakly-supervised training.
    \item Investigating closer integration between the texture prior and base network for possible gains in optimization dynamics.
\end{itemize}